\documentclass[pmlr]{jmlr}% new name PMLR (Proceedings of Machine Learning Research)

\usepackage{longtable}% for long tables

\usepackage{booktabs}
\usepackage[load-configurations=version-1]{siunitx} % newer version
\theorembodyfont{\upshape}
\theoremheaderfont{\scshape}
\theorempostheader{:}
\theoremsep{\newline}

\jmlryear{2022}
\jmlrpages{} % For arXiv
\jmlrworkshop{NeurIPS 2021 Competition and Demonstration Track}

\title[Training Transformers Together]{Training Transformers Together}

\newcommand*\samethanks[1][\value{footnote}]{\footnotemark[#1]}
\author{
  \Name{Alexander Borzunov}\thanks{Equal contribution.}
  \Email{borzunov.alexander@gmail.com}\\
  \Name{Max Ryabinin}\samethanks[1]
  \Email{mryabinin0@gmail.com}\\
  \addr HSE University, Yandex
  \AND
  \Name{Tim Dettmers}\samethanks[1]
  \Email{dettmers@cs.washington.edu}\\
  \addr University of Washington
  \AND
  \Name{Quentin Lhoest}\samethanks[1] \Email{quentin@huggingface.co}\\
  \Name{Lucile Saulnier}\samethanks[1] \Email{lucile@huggingface.co}\\
  \addr Hugging Face
  \AND
  \Name{Michael Diskin} \Email{michael.s.diskin@gmail.com}\\
  \addr HSE University, Yandex
  \AND
  \Name{Yacine Jernite} \Email{yacine@huggingface.co}\\
  \Name{Thomas Wolf} \Email{thomas@huggingface.co}\\
  \addr Hugging Face
}

\begin{document}

\maketitle

\begin{abstract}
The infrastructure necessary for training state-of-the-art models is becoming overly expensive, which makes training such models affordable only to large corporations and institutions. Recent work proposes several methods for training such models collaboratively, i.e., by pooling together hardware from many independent parties and training a shared model over the Internet. In this demonstration, we collaboratively trained a text-to-image transformer similar to OpenAI DALL-E. We invited the viewers to join the ongoing training run, showing them instructions on how to contribute using the available hardware. We explained how to address the engineering challenges associated with such a training run (slow communication, limited memory, uneven performance between devices, and security concerns) and discussed how the viewers can set up collaborative training runs themselves. Finally, we show that the resulting model generates images of reasonable quality on a number of prompts.
\end{abstract}

\section{Introduction}
\label{sec:intro}

Training state-of-the-art deep learning models is becoming ever more computationally demanding. One infamous example of this trend is transformers \citep{transformer}, a popular architecture widely used in NLP \citep{bert,roberta,gpt3}, speech processing \citep{asrtransformer,ttstransformer}, and computer vision \citep{vit,deit,dino}. Transformers benefit from having billions of parameters \citep{gpt3,kaplan2020scaling,ott2018scaling} and large-batch training \citep{Popel2018TrainingTF}, which makes them dependent on large-scale training infrastructure \citep{megatron2,shoeybi2019megatron,Lepikhin2020GShardSG}.

Unfortunately, this kind of infrastructure can be prohibitively expensive, whether one buys the hardware or rents cloud resources \citep{gpt3cost,gpt3costlambda}. As a result, most researchers simply cannot afford to conduct the necessary experiments to develop their ideas, which ultimately slows down scientific progress.

To make large-scale deep learning more accessible, recent work proposes to train these models collaboratively, i.e., to pool together the hardware from many independent parties and train a shared model over the Internet \citep{lc0,hivemind_dmoe,volunteer_dl_async,atre2021distributed,dedloc}. Such work proposes general distributed algorithms for training on many devices with uneven compute capability and reliability. However, to make them practical, one must overcome several engineering challenges, such as slow communication, limited memory, and security concerns.

In this demonstration, we collaboratively trained a text-to-image transformer similar to DALL-E \citep{dalle}. Our contributions are the following:

\begin{itemize}
    \item We modify the DALL-E model, making it suitable for training over the Internet using the method from \citet{dedloc} and the \verb|hivemind| library \citep{hivemind}. We set up the infrastructure for such a training run and publish the training results.
    % \item We developed a PyTorch library \citep{hivemind} for distributed training that addresses challenges associated with training over the Internet. Next, we modified the DALL-E model \citep{dalle} to make it suitable for such a training.
    % \item We set up the infrastructure for training this model collaboratively, as described in \citet{dedloc} (including the bootstrap nodes, the nodes publishing statistics and model checkpoints, the nodes helping to aggregate gradients, and the training dashboard). We maintained the training run and published the training results.
    % It included (a) the nodes serving as the entry points to the network, that also publish the training statistics to Wandb (cite) and the current model checkpoint to HF Model Hub (cite), and (b) the high-throughput nodes that would help aggregate gradients if all volunteers had slow connection speeds.
    \item We provide a webpage\footnote{See \url{https://training-transformers-together.github.io}} explaining how to join the ongoing training run, address challenges related to collaborative training runs (slow communication, low memory budget, support of heterogeneous devices), and set up such a training run by yourself.
    \item We provide an interactive ``calculator'' that shows the memory consumed by different models in case of using various memory-efficiency techniques.
    Also, we present a tutorial on setting up dataset streaming and model compression using the \verb|datasets| and \verb|bitsandbytes| libraries \citep{datasets,bitsandbytes}.
\end{itemize}

\section{Demonstration Contents}

\subsection{Main webpage}

The central part of our demonstration is a webpage where people can explore the demonstration materials. The webpage describes the motivation behind collaborative training projects, the method for efficient training from \citet{dedloc}, and the ongoing collaborative training of our adapted version of DALL-E (see Section~\ref{sec:training_run}). Here, we also show a plot of the training objective and the number of active participants.

Next, we provide instructions on how to join the training run using free cloud providers or their own GPU. This involves (1)~joining a specific Hugging Face organization, where we can authenticate the users and measure their contribution, and (2)~running a Jupyter notebook \citep{jupyter} with the training code. Our intention was that the user can explore our collaborative training environment through active participation while at the same time reading the detailed explanations of how it works. Here, we also provide the link to the interactive dashboard which shows the statistics and the leaderboard of contributors and provides further information about the training run, such as model checkpoints uploaded to the Model Hub, notebooks for inference, and links to the source code.

Then, we proceed to discuss the engineering challenges of collaborative training runs:

\begin{itemize}
    \item \textbf{Communication efficiency.} Most distributed training algorithms are designed for the networks inside HPC clusters with a 10--100 Gbit/s bandwidth. However, typical Internet connections are orders of magnitude slower (10--100 Mbit/s). To make training over the Internet practical, one can reduce the communication costs using large-batch training \citep{lamb}, gradient compression \citep{Dettmers20158BitAF,deepgradientcompression,powersgd,tang20211}, parameter sharing \citep{albert,xue2021go}, and overlapping computation with communication \citep{zerooffload}.
    \item \textbf{Uneven device performance.}
    Traditional data-parallel training waits for the slowest device on every batch. \citet{dedloc} allow the devices to process different numbers of samples for a batch, while keeping the guarantees of synchronous training.
    % Training devices can have vastly uneven performance since each participant brings their own hardware. This makes traditional data-parallel training inefficient since it has to wait for the slowest device on every batch. \citet{dedloc} resolves this bottleneck by allowing each device to process samples at its own pace, while keeping the guarantees of synchronous training. %Early attempts to circumvent this issue use asynchronous training~\cite{recht2011hogwild} to fully utilize each device~\cite{volunteer_dl_async} at the cost of slower convergence~\cite{zhang2015staleness}. \citet{dedloc} resolve this bottleneck, allowing each device to process batches at its own pace.
    \item \textbf{Memory efficiency.} Distributed training requires either storing all parameters and optimizer statistics on each participant, which is challenging in the case of low-end hardware, or using model parallelism which introduces another level of complexity. Fortunately, the first option is often viable if we reduce the memory consumption with 8-bit optimizers \citep{bitsandbytes}, by offloading the statistics to CPU, with gradient checkpointing or parameter sharing \citep{albert,xue2021go}.
    \item \textbf{Dataset streaming.} Participants often cannot store or even download the whole dataset, since datasets used for pretraining transformers may contain hundreds of gigabytes of data. To address that, one can use dataset streaming tools, such as the \verb|datasets| library \citep{datasets}.
    \item \textbf{Security.} Crucially, the participants only exchange tensors and never send code to be executed on each other's computers. Since a malicious participant also could influence the training outcome by sending wrong tensors, we should either authenticate participants, as described in \citet{dedloc}, and/or use gradient aggregation techniques robust to outliers \citep{karimireddy2020learning,gorbunov2021secure}.
\end{itemize}

Finally, we provide a recipe on how to combine all that and set up a new collaborative training run using the \verb|hivemind| library \citep{hivemind}.

\subsection{Memory calculator}

The demonstration webpage includes an interactive ``calculator'' showing the benefits of various memory-efficiency techniques and their combinations. It can compute the consumption of RAM and GPU memory for BERT \citep{bert}, T5 \citep{Raffel2020ExploringTL}, GPT-2 \citep{radford2019language}, GPT-3 \citep{gpt3}, GPT-J \citep{gpt-j}, and DALL-E \citep{dalle} in case of using 8-bit optimizers, offloading the optimizer statistics to CPU, using gradient checkpointing and parameter sharing.

\subsection{Tutorial on memory-efficiency techniques}

The demonstration webpage refers to a tutorial on setting up dataset streaming with the \verb|datasets| library \citep{datasets} and model compression with the \verb|bitsandbytes| library \citep{bitsandbytes}. The goal of the tutorial is to fine-tune the GPT-2 Large model \citep{radford2019language} on the C4 dataset \citep{Raffel2020ExploringTL} using only a low-end GPU, which is possible with the 8-bit Adam optimizer.

\section{Collaborative Training Run}
\label{sec:training_run}

\subsection{Model}

For the practical example of a collaborative training run, we chose to train a text-to-image transformer similar to DALL-E \citep{dalle}, based on the code from \citet{dalle-pytorch}. Specifically, we used a decoder-only transformer with 1024 hidden units and 64 layers, each of which uses 16 attention heads with a per-head state size of 64 ($\approx$1.1B parameters in total). We alternated the attention masks as in the original paper, i.e., repeated ``row, column, row, row'' masks until the last layer, which had the convolutional mask.

To improve communication and memory efficiency, we tied weights of all ``row, column, row, row'' layer groups \citep{albert} and tied the input and output embeddings \citep{press2016using}, so the model uses $\approx$8x fewer parameters (but the same amount of compute). We also used reversible layers \citep{reversible} to reduce memory usage and rotary embeddings \citep{rotary} to improve training stability.

We replaced dVAE with VQ-GAN \citep{vqgan}, since it has a smaller reconstruction error. We used the checkpoint with $f{=}8$ and the codebook size 8192. Finally, we used CLIP ViT/B-32 \citep{clip} to choose the best 4 out of 128 generated images.

\subsection{Dataset}

We trained the model on the first 100 million image-text pairs from LAION-400M \citep{laion}.
We skipped $\approx$10\% images due to short captions, extreme aspect ratios, and NSFW labels.

Before training, we preprocessed all images with VQGAN and uploaded the VQGAN codes and captions, both compressed with Brotli \citep{brotli}, to the Hugging Face Dataset Hub \citep{datasets}. During training, we streamed the compressed codes instead of the original images, thus consuming $\approx$18x less bandwidth.

\subsection{Training procedure}

We followed the distributed training procedure from \citet{dedloc} and used the 8-bit LAMB optimizer \citep{lamb,bitsandbytes} offloaded to CPU.
We used the linear training schedule with 31250 steps (the first 10\% is the warm-up) and the peak learning rate of $2.5 \cdot 10^{-3}$.
% As in the original paper, $\beta_1 = 0.9$, $\beta_2 = 0.96$, and the weight decay multiplier is $4.5 \cdot 10^{-2}$. We clipped gradients whose norm exceeds 4.
While exchanging gradients and parameters, we used the 8-bit quantization \citep{Dettmers20158BitAF} for tensors with $\ge 2^{16}$ elements and the 16-bit precision for other tensors. Unlike the original paper, we did not use PowerSGD \citep{powersgd}.

\subsection{Results}

The training run lasted for 2.5 months and passed $\approx$80\% of the training schedule. Besides the authors, 37 volunteers have contributed for at least 10 minutes~(see Appendix~\ref{apd:top_volunteers}).

During inference, we note that limiting sampling to top 256 logits or top logits whose probability sums up to $p = 0.75$ greatly improves the image quality. The final model generates realistic images for some prompts but fails to draw correct shapes for the others, while using the appropriate image style, textures, and colors~(see Appendix~\ref{apd:inference_results}). We attribute that to the fact that our model is too small to remember the full diversity of images in LAION-400M. Still, the model can generalize to the concepts not present in the dataset.

\bibliography{pmlr-sample}

\newpage
\appendix

\section{Top Volunteers by Contributed Compute Time}\label{apd:top_volunteers}

\begin{figure}[htbp]
  \centering
  {\vskip -5px}
  {\includegraphics[width=0.5\linewidth]{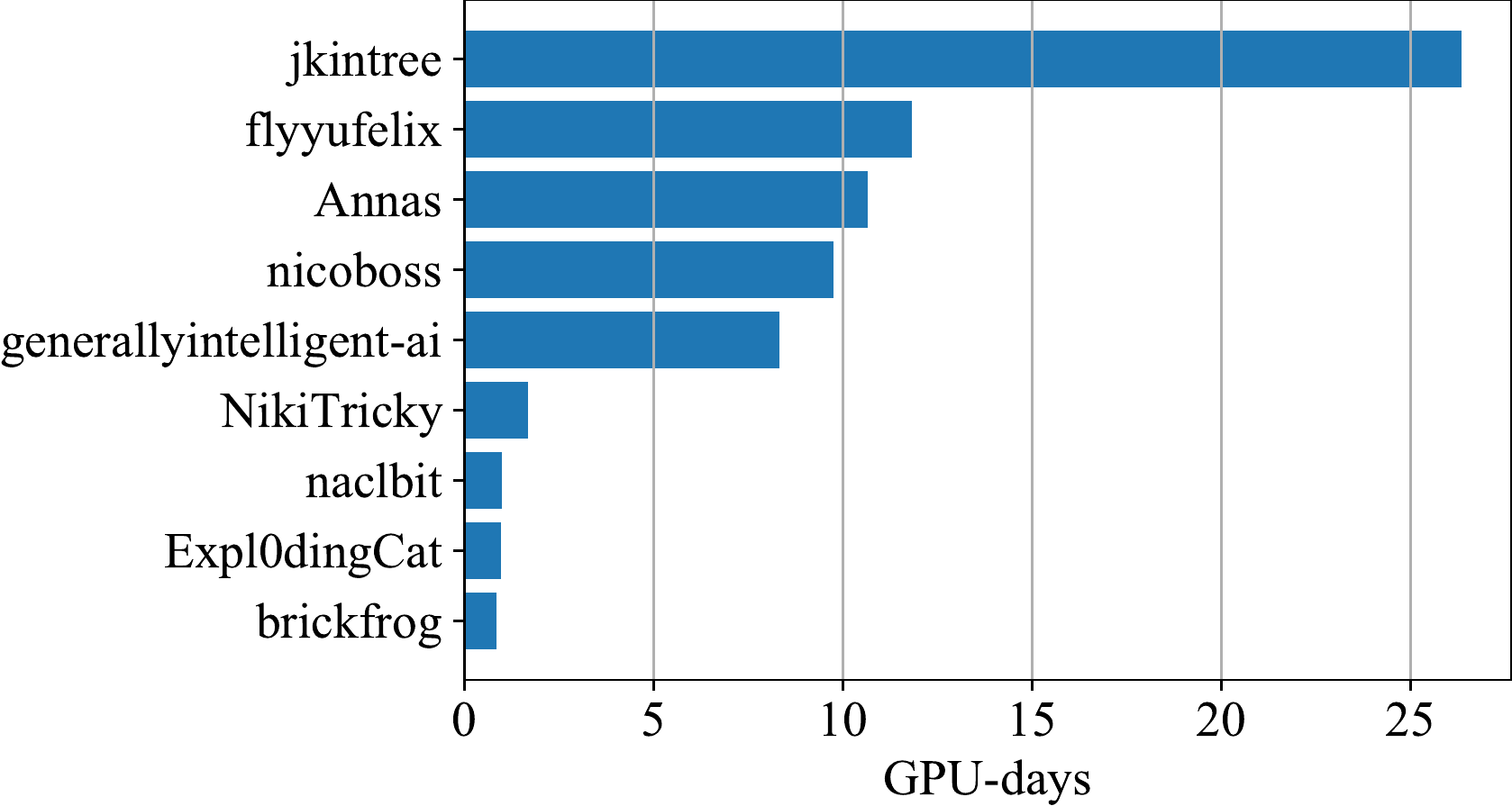}}
  {\vskip -5px}
  {\caption{Hugging Face usernames of volunteers who contributed the most compute time.}}
  {\vskip -10px}
\end{figure}

\section{Model Inference Results}\label{apd:inference_results}

\begin{figure}[htbp]
  \centering
  {\vskip -5px}
  {\includegraphics[width=0.9\linewidth]{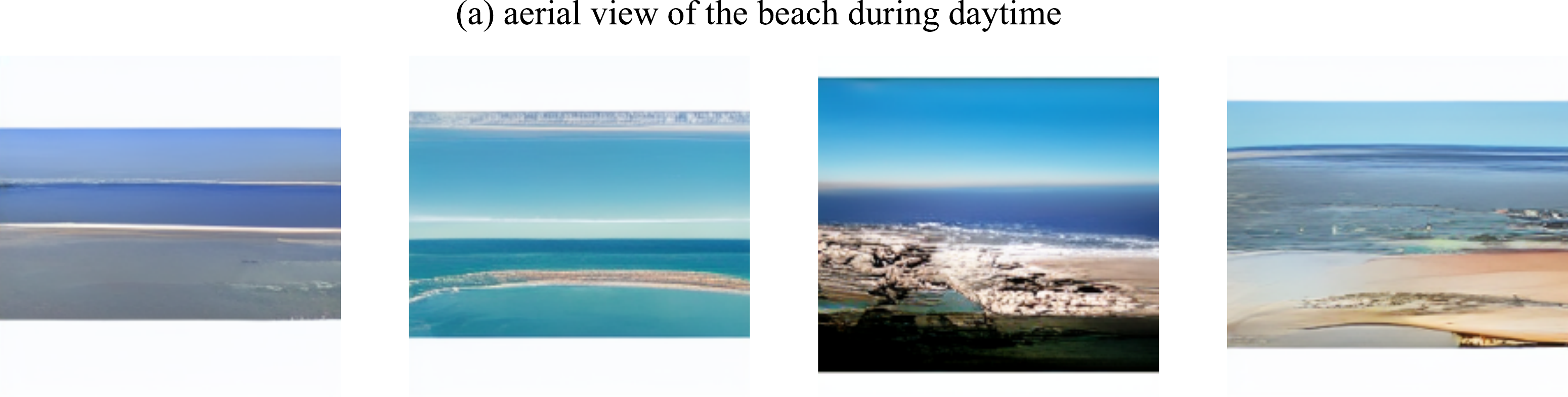}}
  {\includegraphics[width=0.9\linewidth]{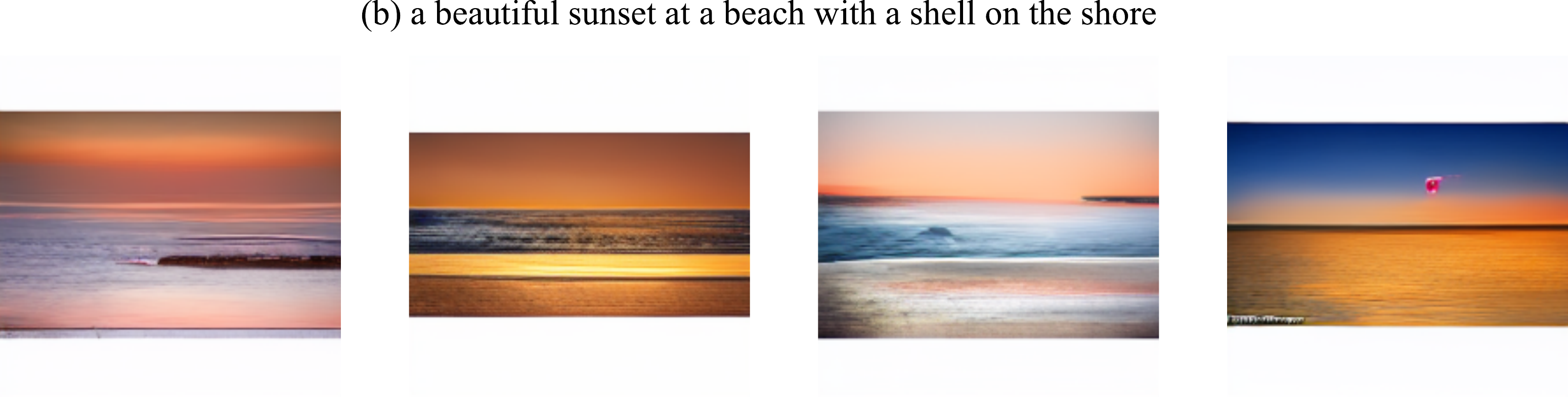}}
  {\includegraphics[width=0.9\linewidth]{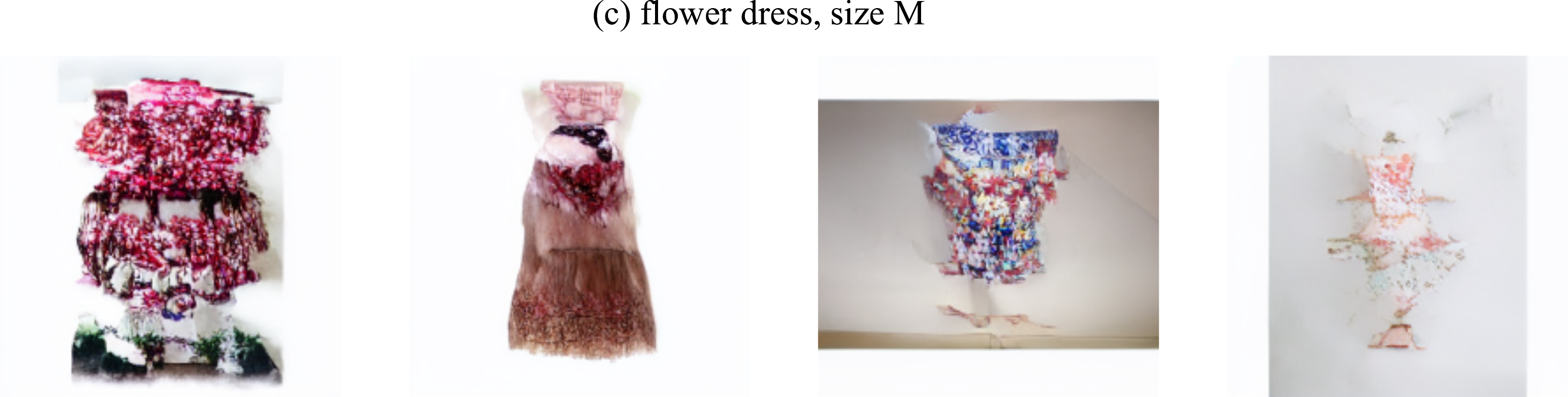}}
  {\includegraphics[width=0.9\linewidth]{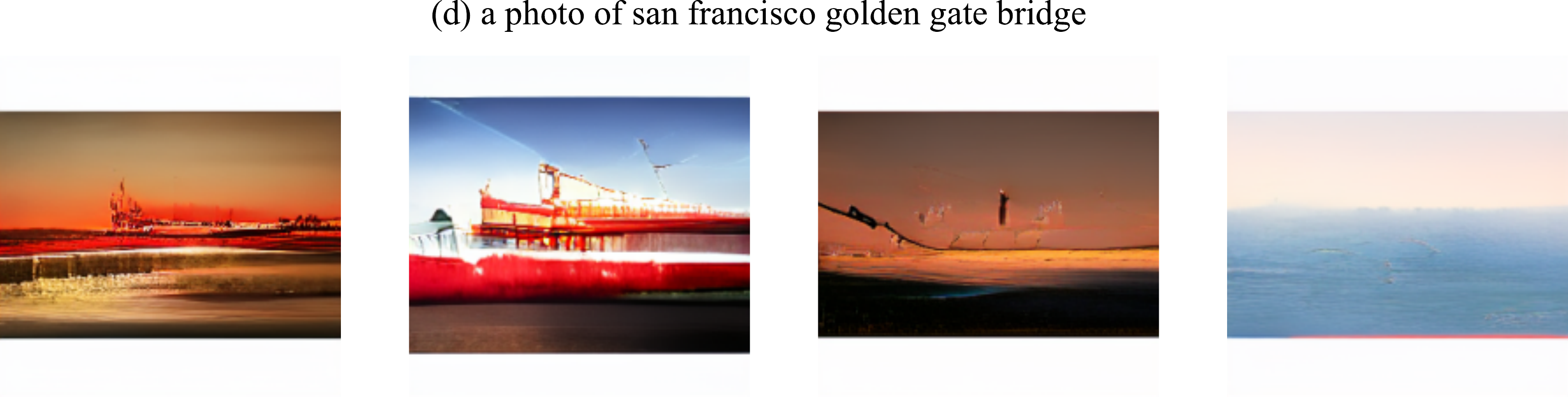}}
  {\vskip -5px}
\end{figure}

\begin{figure}[htbp]
  \centering
  {\includegraphics[width=0.9\linewidth]{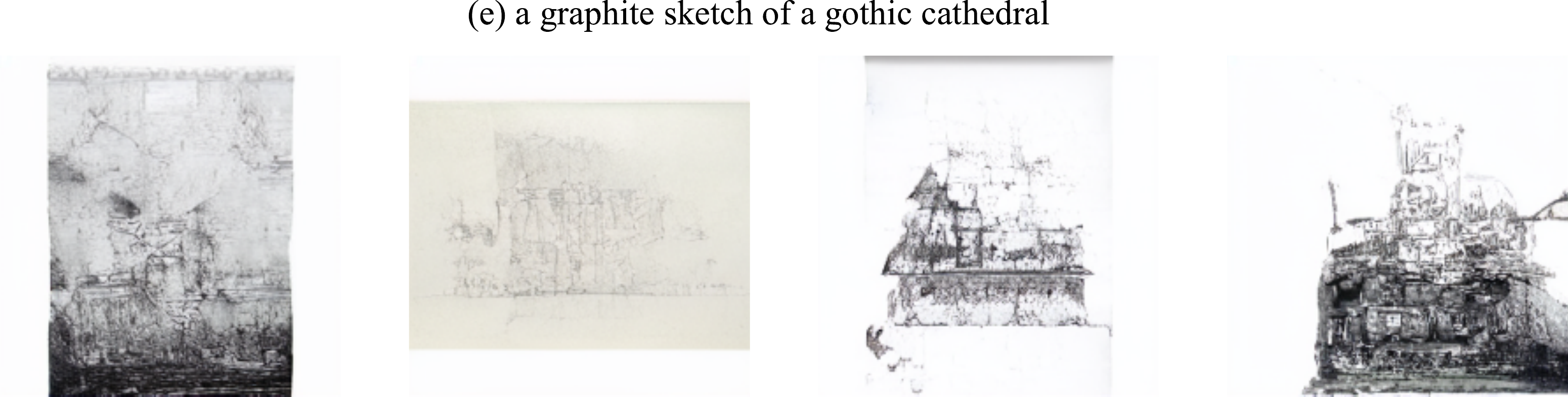}}
  {\includegraphics[width=0.9\linewidth]{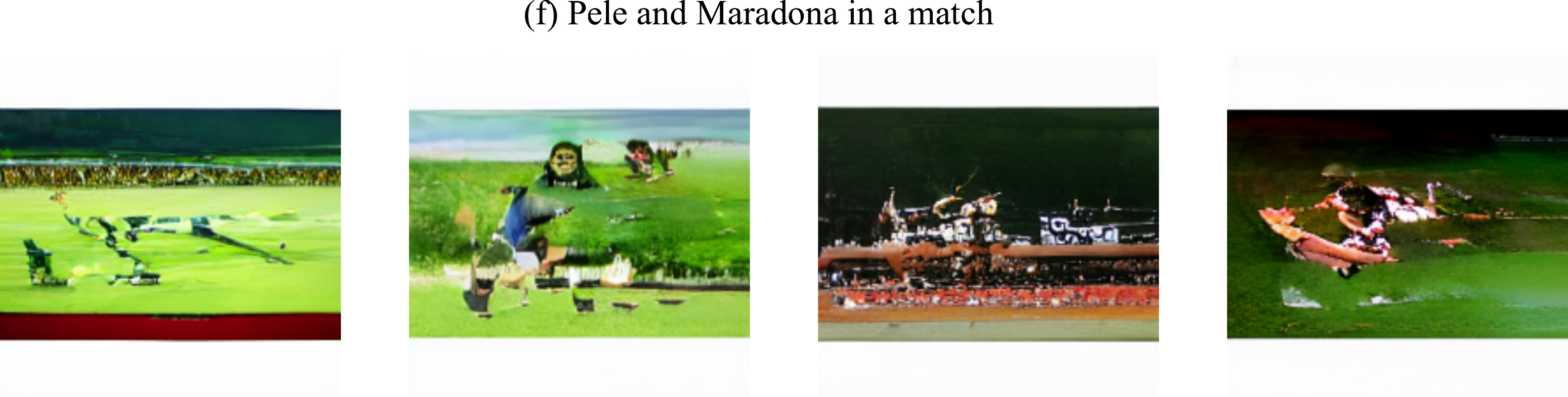}}
  {\includegraphics[width=0.9\linewidth]{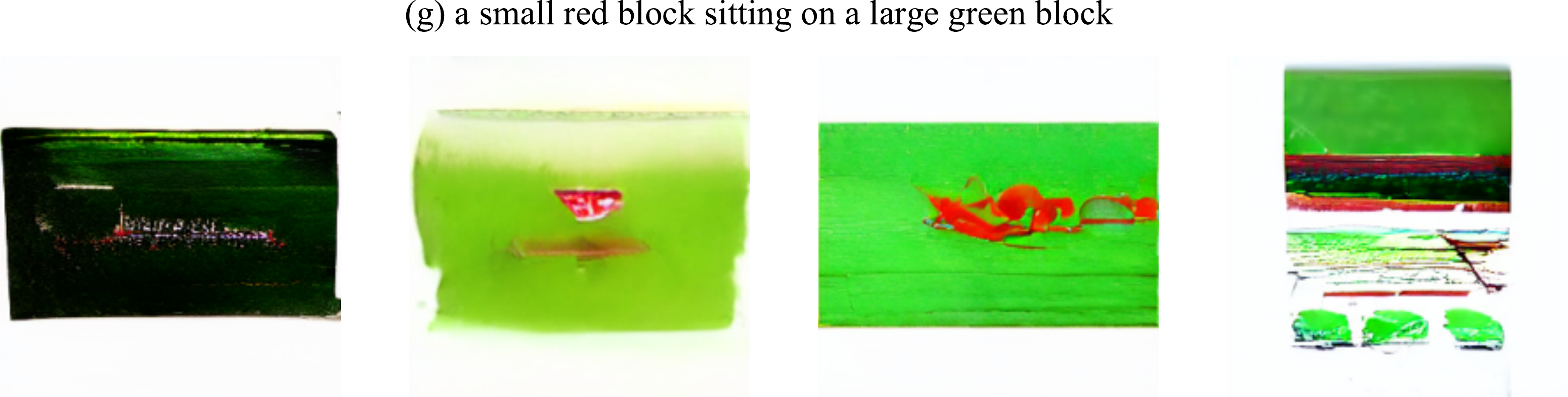}}
  {\includegraphics[width=0.9\linewidth]{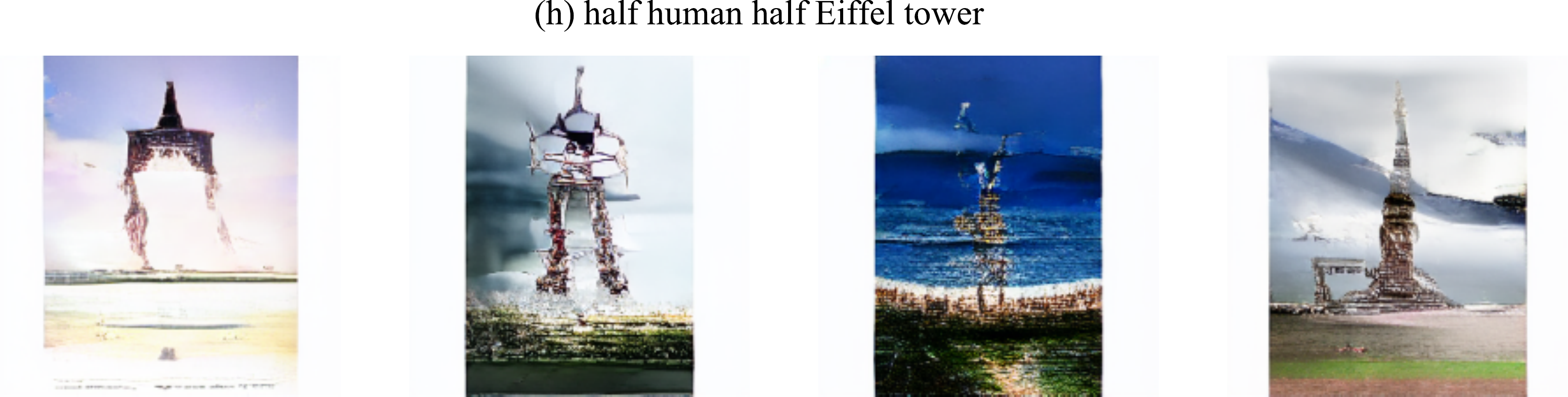}}
  {\vskip 10px}
  {\includegraphics[width=0.9\linewidth]{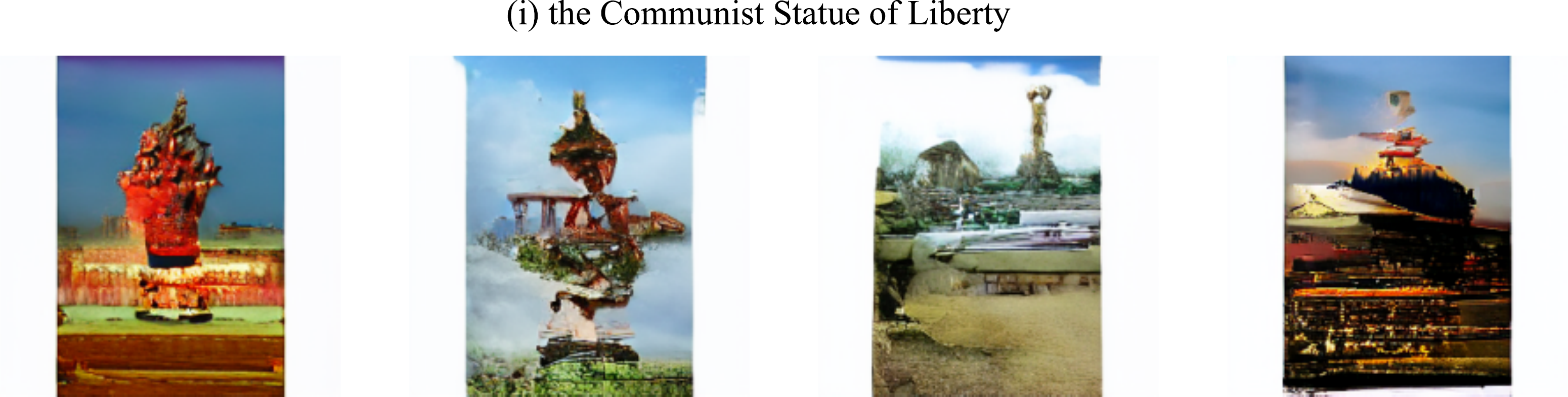}}
  {\caption{Inference results of the final model (the prompts are taken from \citet{dalle-mega-sample-predictions}):\vspace{3px}\\
  \textbf{(a)--(c)} Prompts leading to realistic outputs.\vspace{3px}\\
  \textbf{(d)--(f)} Prompts where the model fails to draw the correct object shapes, but uses the appropriate image style, textures, and colors.\vspace{3px}\\
  \textbf{(g)--(i)} Prompts where the model is able to generalize and draw the concepts not present in the training set. This is checked by inspecting training set images whose CLIP embeddings are close to the prompt embeddings \citep{clip-retrieval}.
  }}
\end{figure}
{\vskip -10px}

\end{document}